\documentclass[10pt,twocolumn,letterpaper]{article}

\usepackage[review]{cvpr}      

\usepackage{graphicx}
\usepackage{amsmath}
\usepackage{amssymb}
\usepackage{booktabs}

\usepackage{cite}

\usepackage{algorithmic}
\usepackage{textcomp}
\usepackage{xcolor}
\usepackage{booktabs}
\usepackage{epstopdf}
\usepackage{makecell}
\usepackage{multirow}

\usepackage[pagebackref,breaklinks,colorlinks]{hyperref}

\usepackage[capitalize]{cleveref}
\Crefname{section}{Sec.}{Secs.}
\Crefname{section}{Section}{Sections}
\Crefname{table}{Table}{Tables}
\Crefname{table}{Tab.}{Tabs.}

\begin{document}
\title{SAT: Size-Aware Transformer for 3D Point Cloud Semantic Segmentation}

\author{Junjie Zhou\,\textsuperscript{\rm 1}, Yongping Xiong\,\textsuperscript{\rm 1}\,\thanks{Corresponding authors.}~, Chinwai Chiu\,\textsuperscript{\rm 1}, Fangyu Liu\,\textsuperscript{\rm 1}, Xiangyang Gong\,\textsuperscript{\rm 1}\\
 \\
\textsuperscript{\rm 1}\,School of Computer Science, Beijing University of Posts and Telecommunications\\
{\tt\small
\{zhoujunjie, ypxiong, chiuchinwai, lfyu, xygong\}@bupt.edu.cn\ 
}
}
\maketitle
\begin{abstract}

Transformer models have achieved promising performances in point cloud segmentation. However, most existing attention schemes provide the same feature learning paradigm for all points equally and overlook the enormous difference in size among scene objects. In this paper, we propose the \textbf{Size-Aware Transformer (SAT)} that can tailor effective receptive fields for objects of different sizes. 
Our SAT achieves size-aware learning via two steps: introduce multi-scale features to each attention layer and allow each point to choose its attentive fields adaptively. It contains two key designs: the Multi-Granularity Attention (MGA) scheme and the Re-Attention module.
The MGA addresses two challenges: efficiently aggregating tokens from distant areas and preserving multi-scale features within one attention layer. Specifically, point-voxel cross attention is proposed to address the first challenge, and the shunted strategy based on the standard multi-head self attention is applied to solve the second.  
The Re-Attention module dynamically adjusts the attention scores to the fine- and coarse-grained features output by MGA for each point.
Extensive experimental results demonstrate that SAT achieves state-of-the-art performances on S3DIS\cite{datasets3dis} and ScanNetV2\cite{datasetscannet} datasets. Our SAT also achieves the most balanced performance on categories among all referred methods, which illustrates the superiority of modelling categories of different sizes.
Our code and model will be released after the acceptance of this paper.
\end{abstract}

\section{Introduction}
\begin{figure}
\centering
\includegraphics[scale=0.3]{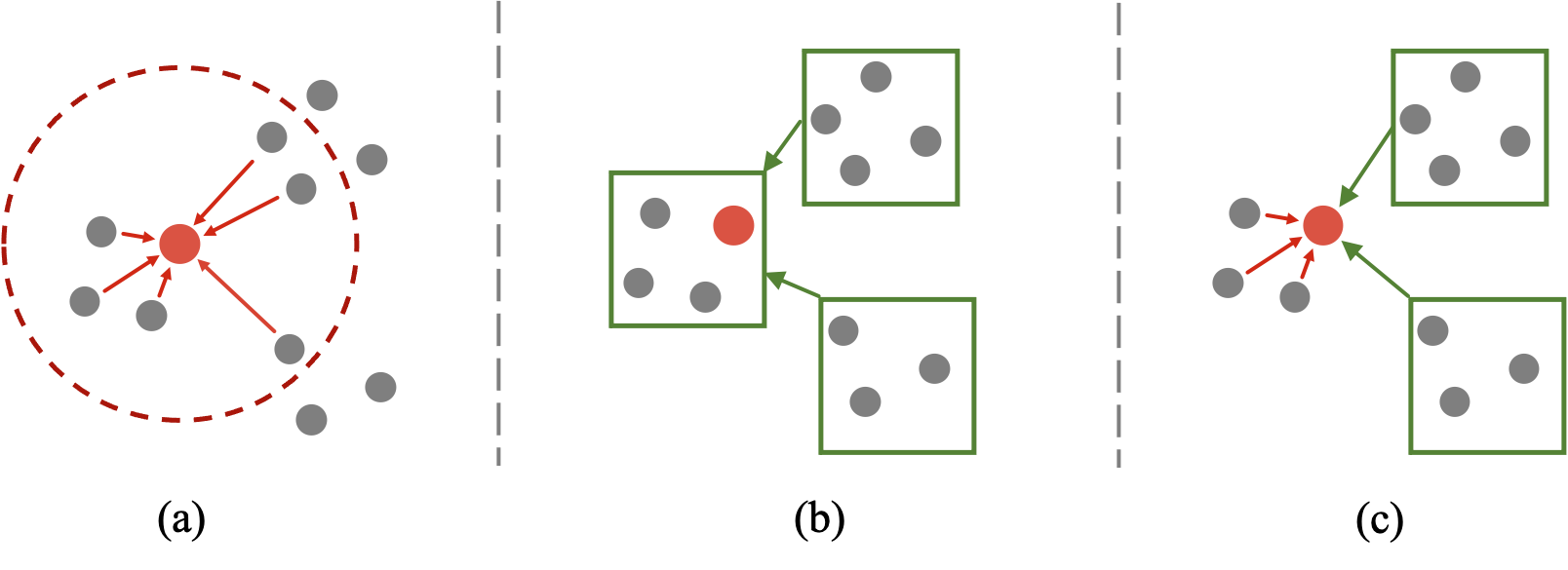}
\caption{The visual comparison of our method with others. \textbf{Left}: point attention. \textbf{Middle}: voxel attention. \textbf{Right}: the proposed multi-granularity attention. Our MGA generates multi-granularity tokens to inject multi-scale features into each attention layer. Unlike (b) voxel attention, we directly apply cross attention between point and voxels without any extra devoxelize operator.}

\label{fig_comparison}
\end{figure}

Semantic segmentation of the 3D point cloud is significant for various applications, such as autonomous driving and robotics. Due to the sparse distribution and irregular representation of the point cloud, mature networks used in images, such as convolution networks, cannot directly process the point cloud. 
Self-attention (SA) is an inherently permutation invariant operator and is fabulous for point cloud processing. Several works \cite{pt2021,pct2021,dtnet2021,strTransformer} make a profound study in transformer architecture\cite{transformer2017} for point cloud learning, and PT\cite{pt2021} surpasses the prior approaches\cite{pointnet, pointnet2, pccn, pointconv, kpconv, randlanet} in scene semantic segmentation by a large margin. It shows that transformer has excellent potential for point cloud segmentation. 

However, the above works\cite{pt2021,pct2021,dtnet2021,strTransformer} take the raw points as input tokens. Due to the computational complexity of SA being quadratic w.r.t. the number of tokens, they are hard to model long-range dependencies for the large scene point cloud. 
Therefore, some follow-up works studied how to inject global features efficiently and introduce multi-scale features to the point transformer.
\cite{pvt2021,crossscaletrans2022} utilize voxel attention that computes the attention maps between voxels to gain more expansive receptive fields. Nevertheless, they must project voxel features to point features via a devoxelize operator, which causes feature information loss. 
PatchFormer\cite{patchformer2022} notices that the size of scene objects varies. It designs multi-scale attention based on voxel attention to build relationships among different size objects. It is efficient enough but detracts from effectiveness due to the lack of direct interaction with fine-grained point features. 

Besides, the prior transformer models fail to achieve size-aware feature learning for the point cloud. \cite{patchformer2022} introduces multi-scale features to voxels, but features of different scales are aggregated as a single entity. We think that objects of various sizes need to focus on different scales. However, the multi-scale features in the above methods are difficult to distinguish for different size objects.

To resolve the above limitations, we develop a novel transformer block, termed \textbf{\textit{Size-Aware Transfomer Block}}, which enables different size objects to adjust their attentive fields adaptively. The block consists of two key components: the Multi-Granularity Attention (MGA) and the Re-Attention module. The MGA extract fine- and coarse-grained features for each point simultaneously in each attention layer. The re-attention module refines the output features from MGA for different size objects. 

The MGA first generates multi-granularity tokens from receptive fields of different sizes. Then, it injects these tokens into each attention layer to preserve multi-granularity features. The MGA addresses two challenges: efficiently aggregates tokens from distant areas and preserves multi-scale features within one attention layer. For the first challenge, we propose point-voxel cross attention to compute attention maps between points and voxels directly. Different from the voxel attention in \cite{pvt2021,crossscaletrans2022,patchformer2022}, which outputs voxel features and further obtains point features by devoxelization, the output of MGA is point features without any additional operator. ~\Cref{fig_comparison} illustrates the differences between our approach and the other two methods. Point attention (a) captures fine-grained features, but the receptive field is limited. Voxel attention (b) acquires large receptive fields, but it is not ideal for point-level prediction tasks because it requires devoxelization. For the second challenge, inspired by\cite{shunted2d2022}, we utilize standard multi-head self-attention and split multiple attention heads into several groups. The multi-granularity tokens from different receptive field sizes are assigned to different groups, hence disentangling the multi-scale features in the learning process for objects. 
Therefore, the MGA provides multi-scale features for multiple-sized objects effectively and efficiently.

Moreover, providing a consistent receptive field for different size objects is not desirable. The re-attention module tailors the weights of different attention heads for each point, which enables different size objects to adjust their attentive fields dynamically. To the best of our knowledge, this is the first time to achieve size-aware feature learning in point cloud segmentation.

We construct our \textbf{Size-Aware Transformer (SAT)} by stacking multiple SAT blocks for point cloud semantic segmentation. Our SAT model achieves state-of-the-art performance on two challenging S3DIS\cite{datasets3dis} and ScanNetV2\cite{datasetscannet} datasets. Notably, SAT achieves the highest IoU on S3DIS in many complex categories compared to the most recent state-of-the-art. SAT also achieves the most balanced results in all categories according to the variance of the category scores.

In total, our contributions are:
\begin{itemize}

    \item We propose the novel Multi-Granularity Attention (MGA) scheme that learns effective multi-scale features within one attention layer for multiple-sized objects.
    \item We propose the Re-Attention module, which could dynamically adjust the attentive fields for objects of different sizes.
    \item We construct our Size-Aware Transformer (SAT) for point cloud segmentation based on the proposed modules. Extensive experimental results in~\Cref{Sec-Exp} demonstrate that SAT achieves state-of-the-art performance.
\end{itemize}

\begin{figure*}
\centering
\includegraphics[scale=0.45]{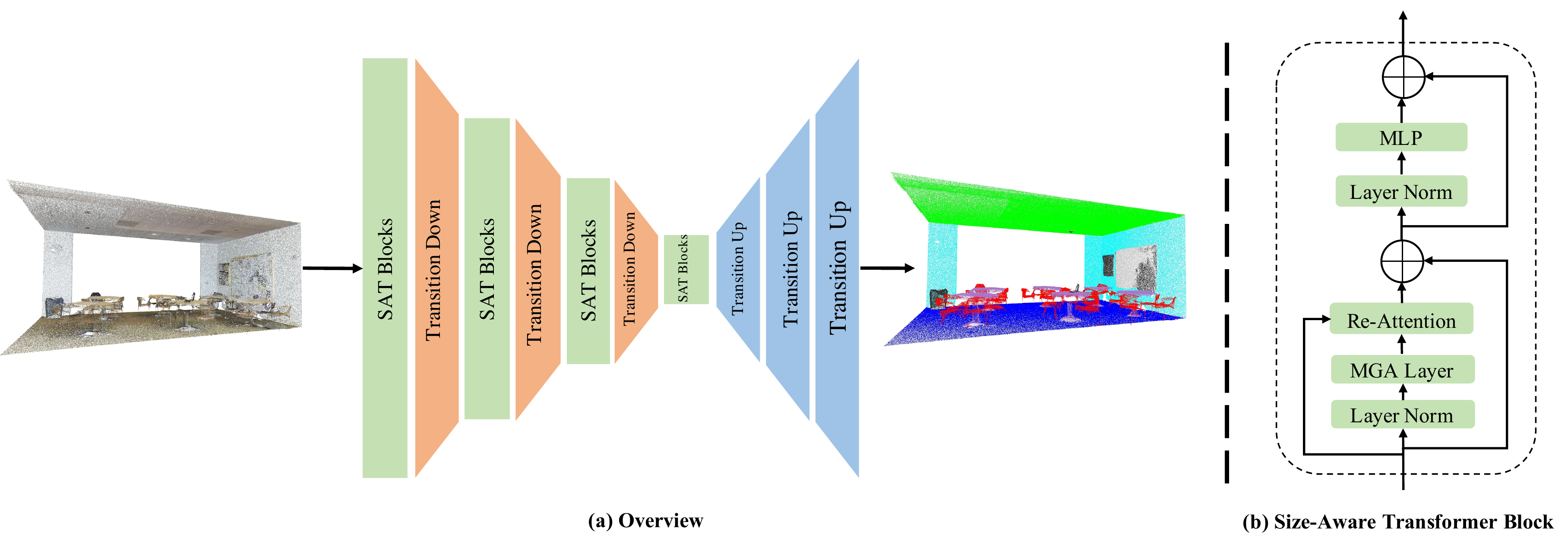}
\caption{(a) The overall architecture of our Size-Aware Transformer. (b) Details of our Size-Aware Transformer block.}
\label{fig-SAT}
\end{figure*}

\section{Related Work}

\subsection{Vision Transformer in 2D Images}
Recently, the pioneering work ViT\cite{ViT2020} and its follow-ups\cite{vittraining2d2021,vittoken2d2021,vitin2d2021,vitDo2d2021} apply a pure attention-based network to image classification and achieve promising performance. Afterwards, \cite{pvt2D2021,swin2d2021} introduce a pyramid structure to broaden the application of Transformer to dense prediction tasks. To reduce the computational cost that increases quadratically with image size, Swin Transformer\cite{swin2d2021} performs local self attention in a fixed-size region and introduces a shifted window mechanism to enable message passing across different local regions. PVT (Pyramid Vision Transformer)\cite{pvt2D2021} designs spatial-reduction attention to merge tokens of \textit{key} and \textit{query}. However, PVT merges too many tokens and thus loses fine-grained information. \cite{shunted2d2022} suggests retaining fine- and coarse-grained information by adaptive merges tokens on large objects and preserving the tokens for small objects. Their proposed shunted self-attention (SSA) achieves the state-of-the-art performance on ADE20K\cite{ADE20K2017} semantic segmentation dataset.

Inspired by SSA\cite{shunted2d2022}, we retain multi-granularity tokens of \textit{key} in our proposed MGA for point cloud segmentation. On top of that, we introduce an efficient re-attention module in our transformer block to enable different size objects to adjust their attentive fields dynamically.
\subsection{Point Cloud Segmentaion}

Motivated by the success of convolution networks in 2D images, many previous works\cite{boulch2017unstructured, SparseConvNet2018, rethage2018fully, vvnet2019} converted the point cloud into discrete voxels and fed it to a 3D convolution network. However, the performance of these methods is sensitive to the granularity of the voxels. For example, coarse-grained voxels have low computational consumption but lower performance due to devoxel operations and lack of fine-grained features.

Since the prior works PointNet/PointNet++\cite{pointnet,pointnet2} directly process 3D data in raw points, extensive works\cite{sonet,pointweb,pccn,pointconv,ecc,gacnet,kpconv,randlanet,scf2021,pcan,boundary-aware2021} introduced permutation invariant neural modules to learn per-point local features. These methods capture fine-grained local features and obtain exciting results. However, most of these works rely on sophisticated kernelization. 

In our work, we combine the advantages of point and voxel representation. Our method efficiently preserves fine- and coarse-grained features without complex kernelization and devoxelize operations.

\subsection{Self-Attention in Point Cloud Segmentation}
Self-attention (SA) is an inherent set operator. Several previous works\cite{pointasnl,randlanet,pointgmm2020} employ SA as an auxiliary module and push the research in point cloud learning. PT\cite{pt2021} and PCT\cite{pct2021} are two pure transformer models for point cloud learning. PT\cite{pt2021} surpasses the concurrent approaches by a large margin, demonstrating the great potential of transformer models in point cloud learning. The follow-up work PCTMA-Net\cite{pctma2021} utilizes standard multi-head attention in point transformer, and Stratified Transformer\cite{strTransformer} introduce shifted window mechanism to point transformer.

However, the above methods are based on point attention, which suffers from limited local receptive fields due to quadratic increase in complexity with input point token. Therefore, many subsequent works\cite{pvt2021,svt2022,patchformer2022,vvattnseg2021,voxeldetr2021, voxeldetr2022} take voxels as tokens and propose the voxel attention that applies self-attention among voxels. These methods always obtain pointwise features via a devoxelize operator related to pointwise prediction. Moreover, PatchFormer\cite{patchformer2022} suggests calculating attention maps between points and patches and directly aggregating patch features for each point. Several of the above methods take the multi-scale feature into account, but none achieve size-aware feature learning.

In our work, we present the point-voxel cross attention in our MGA module. Unlike the voxel attention in previous works, our point-voxel cross attention directly captures relationships between point and voxel without additional devoxelize operators. Moreover, we preserve fine- and coarse-grained tokens within one attention layer and disentangle the multi-scale features in the point cloud learning.

\begin{figure*}
\centering
\includegraphics[scale=0.5]{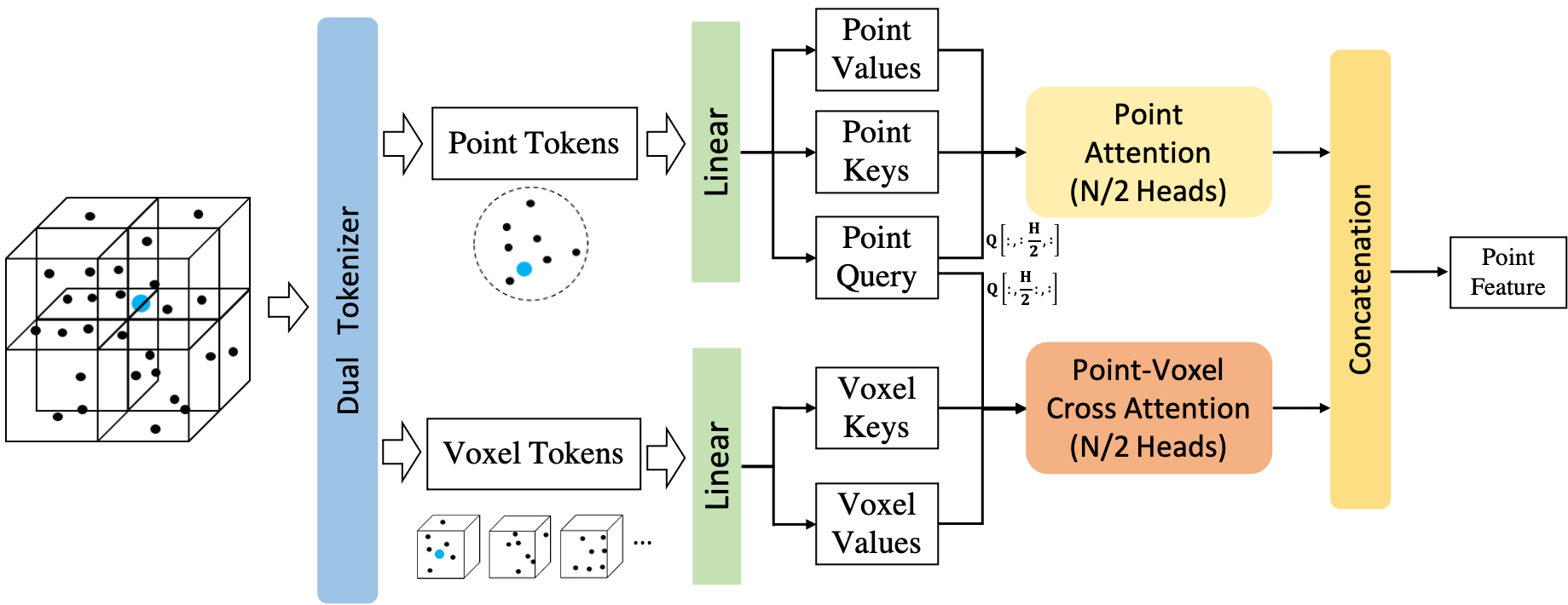}
\caption{The illustration of the Multi-Granularity Attention (MGA). The MGA consists of a point attention branch and a point-voxel cross attention branch. The dual tokenizer tokenizes the point cloud to two sets, the point-tokens set and the voxel-tokens set. Besides, the MGA splits multiple attention heads into several groups and assigns two token sets to different ones. Therefore, the MGA disentangles the multi-scale features in the learning process for objects.}
\label{fig_MGA}
\end{figure*}

\section{Size-Aware Transformer}

\subsection{Overview}
The overall architecture of our Size-Aware Transformer is illustrated in~\Cref{fig-SAT} (a). We follow the hierarchical architecture of~\cite{pointnet2}. Inspired by \cite{swin2d2021,strTransformer}, we compute self-attention locally within non-overlapping windows. Beyond a regular window size, a larger window is additionally used per stage to inject heterogeneous receptive field sizes into tokens. 

Our SAT is built upon the Size-Aware Transformer blocks. The key idea of SAT block is to introduce effective multi-scale features to each attention layer and allow each point to adapt its attentive fields. The details of SAT blocks are shown in~\Cref{fig-SAT} (b). We design a novel Multi-Granularity Attention (MGA) scheme to learn effective multi-scale features for the point cloud. We propose a simple but efficient Re-Attention module to dynamically adjust weights of different granularity features for objects of various sizes.

\subsection{Multi-Granularity Attention}\label{sec_MGA}

The architecture of MGA is shown in~\Cref{fig_MGA}. To learn effective multi-scale features for different size objects within each attention layer, the MGA faces two challenges: how to efficiently aggregate tokens from distant areas and how to disentangle multi-scale features during the point learning process. To address the first challenge, we design a novel point-voxel cross attention. To solve the second challenge, we propose the point-voxel shunted strategy based on the standard multi-head self attention.

\subsubsection{Point-Voxel Cross Attention}\label{sec-pvca}

\quad The tokenizer of the existing point transformer models always partitions the point cloud into either a point set or a voxel set. The tokens of \textit{key} and \textit{query} in these attention schemes are from a single set, such as the point attention and the voxel attention. Due to the quadratic complexity w.r.t the number of input points, the receptive fields of point attention are especially limited to the density of the point cloud. The voxel attention alleviates this problem. However, since segmentation is a dense prediction task, it still requires a devoxel operator that projects voxel features to point features, which causes information loss.

In our work, we propose to apply cross attention to a point set and a voxel set, namely, Point-Voxel Cross Attention (PVCA). As shown in~\Cref{fig_MGA}, PVCA directly computes the attention between \textit{queries} from the point set and \textit{key-value} pairs from the voxel set. PVCA is more flexible because it avoids the extra devoxel operation.

Specifically, we first partition the 3D space into non-overlapping voxel windows. Then, our dual tokenizer tokenizes the point cloud to two sets, i.e. a voxel set $\mathcal{V}$ and a point set $\mathcal{P}$. Given that $V_i$ is a subset of $\mathcal{V}$, which contains voxels in the \textit{i}-th voxel-windows. The voxel feature of $v_m\in\{V_i\}$ is formulated as

\begin{equation}
\begin{split}
& v_m = \boldsymbol{\phi}(\frac{1}{n_k} \sum_{k}p_{mk})\\
\end{split}
\end{equation}
where $v_m$ is the \textit{m}-th voxel feature in set $V_i=\{v_m\}_i$, $p_{mk}$ is the \textit{k}-th point feature in the \textit{m}-th voxel, $n_k$ is the number of points in the \textit{m}-th voxel, function $\boldsymbol{\phi}$ is an MLP with two linear layers and one GELU nonlinearity.

Next, given the point features set $F \in \mathbb{R}^{D_n\times(D_h \times D_d)}$ which is located in the \textit{i}-th voxel-window, where $D_n$ is the number of points located in the \textit{i}-th voxel-window, $(D_h \times D_d)$ is the point feature dimension. Formally, the PVCA in the \textit{i}-th voxel window is calculated as follows:
\begin{equation}
\begin{split}
& v_m, F = {\rm \textbf{LN}}(v_m), {\rm \textbf{LN}}(F)\\
& K_v, V_v = v_m {W}_1^K, v_m {W}_1^V\\
& Q_1, Q_2 = F{W}_1^Q, F{W}_2^Q\\
\end{split}
\end{equation}
\begin{equation}
\begin{split}
& F_{coarse} = {\rm \textbf{MSA}}_{\frac{H}{2}}(Q_1, K_v, V_v)\\
\end{split}    
\end{equation}
where \textbf{LN} is a LayerNorm layer, $W$ here are linear matrices, \textbf{MSA} is the multi-head self attention, and $F_{coarse}\in\mathbb{R}^{D_n\times(\frac{D_H}{2}\times D_d)}$ is the output point feature of PVCA. Note that all the linear matrices $W$ project the input vectors to output vectors with only half the dimension of the inputs.

In the specific implementation, we establish voxel token indexes for each point individually and in parallel. The above steps can be easily achieved with the scatter function\footnote{https://github.com/rusty1s/pytorch\_scatter}.

Notably, we reduce the computational complexity from $O(N_p^2)$ to $O(N_pN_v)$ compared to point attention, where $N_v \ll N_p$. Besides, we halve the \textit{query, key, and value} dimensions through the linear layer, which further reduces the calculation consumption. Our PVCA directly computes attention maps between points and voxels, which can model more precise relations between points and coarse-grained features.

\begin{table*}[]\centering
\begin{tabular}{c|c|c|c|c}
\toprule
        & Output Size & Layer Name              &  S3DIS\cite{datasets3dis} & ScanNetV2\cite{datasetscannet} \\
\midrule
Stage 1 & 48      & \begin{tabular}[c]{@{}c@{}}SAT \\Trans. Block\end{tabular}            &  \begin{tabular}[c]{@{}c@{}}$r_i=\left\{\begin{matrix}1^3, & i\leq \frac{Head}{2}  \\ 2^3, & i>  \frac{Head}{2}  \end{matrix}\right. $\\ BW=0.16, Vox=0.08, N=2 \end{tabular} & \begin{tabular}[c]{@{}c@{}}$r_i=\left\{\begin{matrix}1^3, & i\leq \frac{Head}{2}  \\ 3^3, & i>  \frac{Head}{2}  \end{matrix}\right. $\\ BW=0.1, Vox=0.1, N=3 \end{tabular}       \\
\midrule
Stage 2 & 96  & \begin{tabular}[c]{@{}c@{}}SAT \\Trans. Block\end{tabular} &  \begin{tabular}[c]{@{}c@{}}$r_i=\left\{\begin{matrix}1^3, & i\leq \frac{Head}{2}  \\ 2^3, & i>  \frac{Head}{2}  \end{matrix}\right. $\\ BW=0.32, Vox=0.16, N=2 \end{tabular}      &  \begin{tabular}[c]{@{}c@{}}$r_i=\left\{\begin{matrix}1^3, & i\leq \frac{Head}{2}  \\ 3^3, & i>  \frac{Head}{2}  \end{matrix}\right. $\\ BW=0.2, Vox=0.2, N=6 \end{tabular}        \\
\midrule
Stage 3 & 192     & \begin{tabular}[c]{@{}c@{}}SAT \\Trans. Block\end{tabular} &  \begin{tabular}[c]{@{}c@{}}$r_i=\left\{\begin{matrix}1^3, & i\leq \frac{Head}{2}  \\ 2^3, & i>  \frac{Head}{2}  \end{matrix}\right. $\\ BW=0.64, Vox=0.16, N=6 \end{tabular}      &  \begin{tabular}[c]{@{}c@{}}$r_i=\left\{\begin{matrix}1^3, & i\leq \frac{Head}{2}  \\ 3^3, & i>  \frac{Head}{2}  \end{matrix}\right. $\\ BW=0.4, Vox=0.2, N=6 \end{tabular}       \\
\midrule
Stage 4 & 384     & \begin{tabular}[c]{@{}c@{}}SAT \\Trans. Block\end{tabular}         &  \begin{tabular}[c]{@{}c@{}}$r_i=\left\{\begin{matrix}1^3, & i\leq \frac{Head}{2}  \\ 2^3, & i>  \frac{Head}{2}  \end{matrix}\right. $\\ BW=1.28, Vox=0.32, N=2 \end{tabular}      &  \begin{tabular}[c]{@{}c@{}}$r_i=\left\{\begin{matrix}1^3, & i\leq \frac{Head}{2}  \\ 3^3, & i>  \frac{Head}{2}  \end{matrix}\right. $\\ BW=0.8, Vox=0.4, N=6 \end{tabular}        \\
\midrule
Stage 5 & 384     & \begin{tabular}[c]{@{}c@{}}SAT \\Trans. Block\end{tabular}        & -     &  \begin{tabular}[c]{@{}c@{}}$r_i=\left\{\begin{matrix}1^3, & i\leq \frac{Head}{2}  \\ 3^3, & i>  \frac{Head}{2}  \end{matrix}\right. $\\ BW=1.6, Vox=0.4, N=3 \end{tabular}             \\
\bottomrule
\end{tabular}
\caption{Detailed architecture specifications. BW represent the base window size of each stage, $r_i$ is the ratio of point-window or voxel-window size to the the BW. N indicates the number of blocks per stage. Vox is the voxel size in PVCA.}
\label{tbl-modelarch}
\end{table*}

\subsubsection{Point-Voxel Shunted Strategy} \label{sec-pvshunt}
\paragraph{Two Branches of MGA.}As shown in~\Cref{fig_MGA}, our MGA contains the point attention branch and the PVCA branch. For the PVCA branch, it models coarse-grained features in the large field. For the point attention branch, due to its expensive computational cost, we restrict it to computing within a smaller local region, which helps reduce computational consumption.

\paragraph{Point-Voxel Shunted Strategy.}Inspired by \cite{shunted2d2022}, our MGA applies the multi-head self-attention mechanism to combine the multi-scale features from the PVCA branch and the point attention branch. Formally, the MGA are computed as
\begin{equation}
\begin{split}\label{equa-MGA}
& K_p, V_p = F_p {W}_2^K, F_p {W}_2^V\\
& F_{fine} = {\rm \textbf{MSA}}_{\frac{H}{2}}(Q_2, K_p, V_p) \\
& F^{'} = {\rm \textbf{Concat}}(F_{coarse}, F_{fine})
\end{split}
\end{equation}
where all the $W$ here are linear matrices that halves the input vector dimensions. $F^{'} \in \mathbb{R}^{D_n\times(D_h\times D_d)}$ is the final output feature of MGA layer,

In this way, our transformer block can learn multi-scale features within one attention layer in a disentangling way. By stacking several size-aware transformer blocks, the SAT can combine a range of receptive field sizes from different hierarchical layers. 

We emphasize again that MGA is efficient even though it has two branches because we restrict the set base of \textit{point keys} and halve the dimensions of each branch.

\begin{table*}[h]\footnotesize 
\centering
\begin{tabular}{{l|cc|ccccccccccccc}}
\toprule
Methods  & \makecell[c]{\textbf{mIoU}\\(\%)} &\makecell[c]{\textbf{mAcc}\\(\%)}  & \rotatebox{90}{Ceil.}  & \rotatebox{90}{Floor} & \rotatebox{90}{Wall} & \rotatebox{90}{Beam}  & \rotatebox{90}{Col.}   & \rotatebox{90}{Wind.}& \rotatebox{90}{Door}& \rotatebox{90}{Table}& \rotatebox{90}{Chair}& \rotatebox{90}{Sofa}& \rotatebox{90}{Book.}& \rotatebox{90}{Board}&\rotatebox{90}{Clut.}         \\
\midrule
PointNet\cite{pointnet}               & 41.1 &66.2  & 88.8  & 97.3  & 69.8  & 1.0 & 3.9   & 46.3  & 10.8  & 59.0    & 52.6  & 5.9   & 40.3  & 26.4  & 33.2   \\
RSNet\cite{rsnet2018}               & 51.9 &59.4  & 93.3  & 98.3  & 79.2  & 0.0 & 15.8   & 45.4  & 50.1  & 67.9    & 65.5  & 52.5   & 22.5  & 41.0  & 43.6 \\
PointCNN\cite{pointcnn2018}               & 57.3 &63.9 & 92.3  & 98.2  & 79.4  & 0.0 & 17.6  & 22.8  & 62.1  & 74.4  & 80.6  & 31.7  & 66.7  & 62.1  & 56.7   \\
SPGraph\cite{spgraph2018}                & 58.0  &66.5    & 89.4  & 96.9  & 78.1  & 0.0 & 42.8  & 48.9  & 61.6  & 84.7  & 75.4  & 69.8  & 52.6  & 2.1   & 52.2   \\
PCCN\cite{pccn}                   & 58.3 &67.0 & 92.3  & 96.2  & 75.9  & \textbf{3.0} & 6.0     & \textbf{69.5}  & 63.5  & 66.9  & 65.6  & 47.3  & 68.9  & 59.1  & 46.2   \\
PointWeb\cite{pointweb}               & 60.3 &66.6  & 92.0    & 98.5  & 79.4  & 0.0 & 21.1  & 59.7  & 34.8  & 76.3  & 88.3  & 46.9  & 69.3  & 64.9  & 52.5   \\
MinkowsikiNet\cite{Minkowski2019}          & 65.4  &71.7  & 91.8  & \textbf{98.7}  & 86.2  & 0.0 & 34.1  & 48.9  & 62.4  & 81.6  & 89.8  & 47.2  & 74.9  & 74.4  & 58.6   \\
KPConv\cite{kpconv}                 & 67.1  &72.8  & 92.8  & 97.3  & 82.4  & 0.0 & 23.9  & 58.0    & 69.0    & 81.5  & 91.0    & 75.4  & 75.3  & 66.7  & 58.9   \\
ASSANet-L\cite{assanet2021}              & 66.8  &-   & -     & -     & -     & - & -     & -     & -     & -     & -     & -     & -     & -     & -      \\
RepSurf-U\cite{repsurf2022}              & 68.9  &76.0   & -     & -     & -     & - & -     & -     & -     & -     & -     & -     & -     & -     & -      \\
CBL\cite{cbl2022}                    & 69.4  &75.2   & 93.9  & 98.4  & 84.2  & 0.0 & 37.0    & 57.7  & 71.9  & 91.7  & 81.8  & 77.8  & 75.6  & 69.1  & 62.9   \\

PatchFormer\cite{patchformer2022}            & 68.1  &-   & -     & -     & -     & - & -     & -     & -     & -     & -     & -     & -     & -     & -      \\

Fast PT.\cite{fasttrans2022}  & 70.1 &77.4    & -     & -     & -     & - & -     & -     & -     & -     & -     & -     & -     & -     & -      \\
Point Transformer\cite{pt2021}      & 70.4  &76.5  & 94.0    & 98.5  & 86.3  & 0.0 & 38.0    & 63.4  & 74.3  & 82.4  & 89.1  & 80.2  & 74.3  & 76.0    & 59.3   \\

PointNeXt-XL\cite{pointnext2022}           & 70.8  &77.5   & 94.2  & 98.5  & 84.4  & 0.0 & 37.7  & 59.3  & 74.0    & 83.1  & 91.6  & 77.4  & 76.72 & 78.8  & 60.6   \\
PT v2\cite{ptv22022}                 & 71.6  &77.9   & -     & -     & -     & - & -     & -     & -     & -     & -     & -     & -     & -     & -      \\
Swin3dFormer\cite{strTransformer}                 & 70.2  &-    & -     & -     & -     & - & -     & -     & -     & -     & -     & -     & -     & -     & -      \\
StratifiedFormer\cite{strTransformer} & 72.0   &78.1    & \textbf{96.2} & \textbf{98.7}  & 85.6 & 0.0 & 46.1 & 60.0 & \textbf{76.8} & \textbf{92.6} & 84.5 & 77.8 & 75.2 & 78.1  & \textbf{64.0}  \\

\midrule

\textbf{SAT}             & \textbf{72.6}  &\textbf{78.8}  & 93.6 & 98.5 & \textbf{87.2} & 0.0 & \textbf{49.3} & 61.1 & 73.6 & 83.7  & \textbf{91.8} & \textbf{81.7} & \textbf{77.9} & \textbf{82.3} & 63.4 \\
\bottomrule

\end{tabular}
\caption{Quantitative results on S3DIS\cite{datasets3dis} Area 5 dataset, the classwise metric is IoU (\%).}
\label{tbl-results-s3dis}
\end{table*}

\subsection{Re-Attention Module}

Standard multi-head attention allows the transformer to focus jointly on features from different representation sub-spaces at different perspectives. Therefore, these heads are independent and have non-interaction. 

In our MGA, each attention head has its specific meaning, i.e. different heads represent information of different granularity from local or distant areas. However, different size objects should pay different attention to different granularity features. Intuitively, the small object chairs with much geometric information tend to learn fine-grained features, while the large object walls require large receptive fields to perceive surrounding objects.

Therefore, we propose to tailor the attentive fields for multiple-sized objects and design a simple but effective Re-Attention module to apply distinctive weights to different heads.

As shown in~\Cref{fig-SAT} (b), the re-attention module is applied after the MGA layer. Formally, the re-attention module is computed as:
\begin{equation}
\begin{split}\label{equa-reattention}
& \alpha = Sigmoid(\boldsymbol{\gamma}(F))\\
& F_{out} = \alpha \odot F^{'}
\end{split}
\end{equation}
where $\odot$ is element-wise multiplication, $\boldsymbol{\gamma}$ is an MLP with two linear layers and one GELU nonlinearity, $F$ is the point feature before input into MGA, $F^{'}$ is the point feature output by former MGA layer, and $F_{out}$ is the final output point feature of re-attention module. The hidden dimension of $\boldsymbol{\gamma}$ is consistent with the number of attention heads, thus corresponding to different granularity features.

\section{Experiments}\label{Sec-Exp}
We evaluate our SAT on S3DIS\cite{datasets3dis} and ScanNet\cite{datasetscannet} datasets. We report the comparison results with the current state-of-the-art methods. In addition, we conduct extensive ablation experiments and analyze the impact of our proposed module.

\subsection{Implementation Details and Settings}\label{sec-4.1}
\paragraph{Architecture Settings.} As shown in~\Cref{fig-SAT}, our model is based mainly on proposed MGA Transformer blocks, which mainly consist of the point attention and PVCA branches. Table~\ref{tbl-modelarch} shows the model variants of our methods for different tasks and datasets. BW is the base window size of each stage, and $r_i$ indicates the ratio of the spatial computation range of each branch compared to the base window. Corresponding to~\Cref{fig_MGA}, point attention is assigned to the first half of the head of MGA, while PVCA is the second half. The Voxel Window is always larger than the point attention. For deeper stages, we will correspondingly enlarge the voxel size. The down-sample ratio of transition down layers are set to 4.

\paragraph{Training Strategies.} Unless expressly specified, the initial learning rate is set to 0.006 and is dropped with the multiple-step decay. In the case of the S3DIS dataset, we train for 100 epochs with 2 RTX 3090 GPUs. For the ScanNetV2 dataset, we train for 120 epochs with 4 RTX 3090 GPUs. Each point is represented by 3D coordinates and color information. For fairly evaluating the model performance, we apply cross-entropy loss for training and do not use any pre-training strategy.

\begin{figure*}
\centering
\includegraphics[scale=0.6]{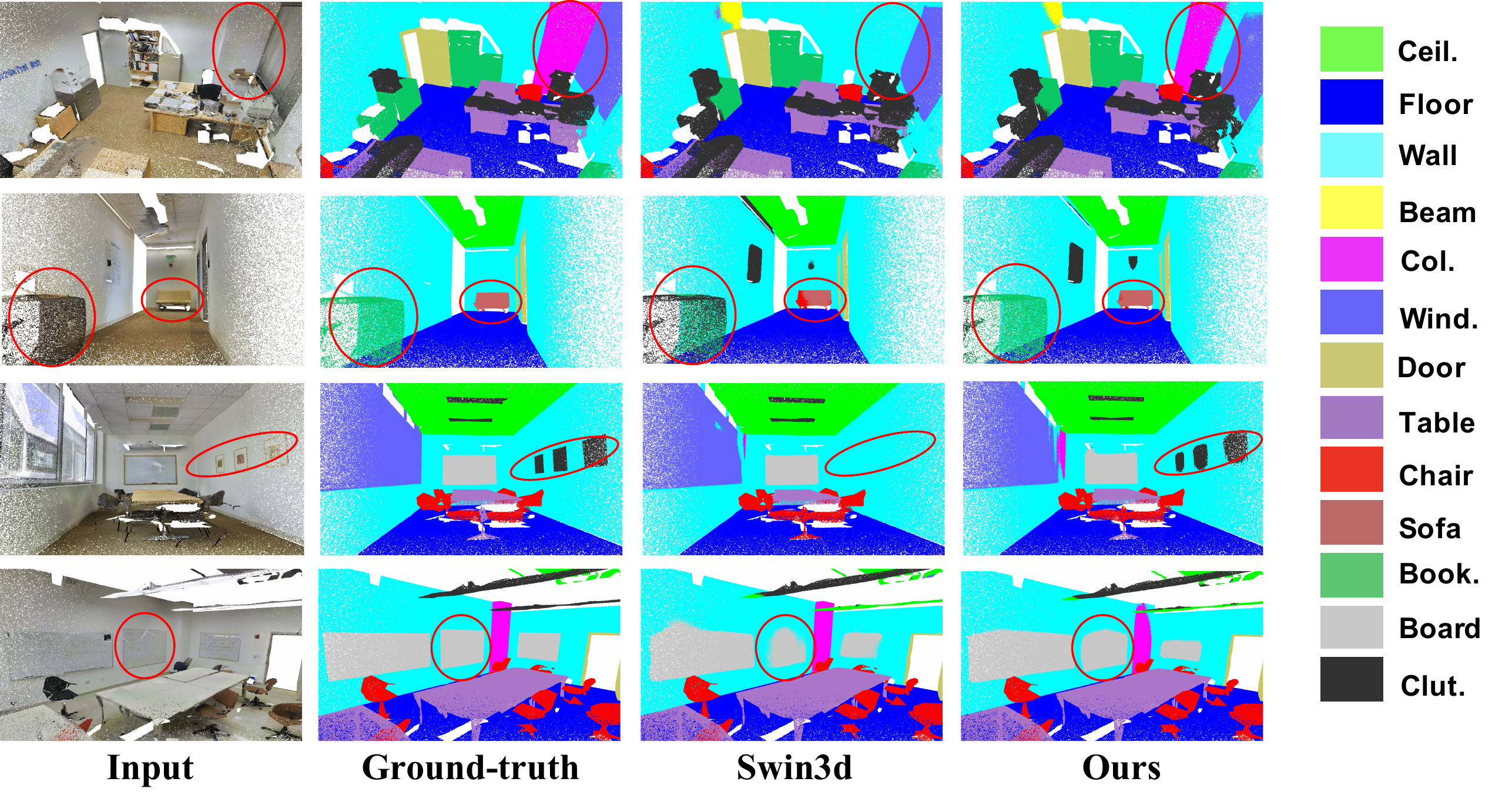}
\caption{Visualization examples of some typical indoor scenes on S3DIS. SAT performs well in many difficulty categories (such as column, sofa, and board). Due to the size-aware capability of our SAT, it is able to maintain stable performance on both large and small objects.}
\label{Fig-vis-s3dis}
\end{figure*}

\subsection{Results}\label{sec-4.2}
\subsubsection{Evaluation on S3DIS Dataset}

\quad S3DIS\cite{datasets3dis} is a challenging benchmark composed of 6 large-scale indoor areas, 271 rooms, and 13 semantic categories. Following the common protocol\cite{segcloud2017}, we evaluate the presented methods in Area 5 and train on the rest.

We report the quantitative results of the current state-of-the-art methods and ours in Table~\ref{tbl-results-s3dis}. The evaluation metrics are the mean Intersection-over-Union (mIoU) and the mean class Accuracy (mAcc). SAT achieves new state-of-the-art performance with the highest value on mIoU and mAcc, and performs better than others in 6 categories. Moreover, we calculated the variance of all reported methods on class-wise IoU to measure the stability of objects of different sizes. The variance of SAT is the lowest, which indicates that SAT has the most robust modelling capability on different size objects.

\Cref{Fig-vis-s3dis} shows the visualization examples of our method in Area 5. SAT works well to capture complex interior architectural structures, such as the hard-to-distinguish columns. In addition, distinguishing small objects from large building constructions is quite difficult because it requires a strong ability to model relationships among objects of different sizes. Due to the power of dynamically adjusting attentive fields for different size objects, SAT can distinguish small paintings (clutters) on large walls and has better boundary segmentation performance on boards.

\subsubsection{Evaluation on ScanNetV2 Dataset}
\quad ScanNetV2\cite{datasetscannet} consists of 1631 indoor scans, including 1201 training scenes, 312 validation scenes, and 100 test scenes. Each point in the training and validation set is labelled in 20 categories.

We compare the mIoU of our method on the validation and test set in Table~\ref{tbl-scannet}. SAT outperforms the prior methods in both the validation and testing sets, with 74.4\% mIoU and 74.2\%, respectively.

\begin{table}
\begin{center}
\begin{tabular}{l|c|c}
\toprule
Methods                         & \makecell[c]{Val mIoU \\(\%)}                    &\makecell[c]{Test mIoU \\(\%)} \\ 
\midrule
PointNet++\cite{pointnet2}           & -            & 33.9 \\ 
PointEdge\cite{pointedge2019}            & 63.4             & 61.8 \\ 
PointConv\cite{pointconv}            & 61.0             & 66.6 \\ 
PointASNL\cite{pointasnl} & 66.4             & 66.6 \\ 
BA-GEM\cite{boundary-aware2021} &     -       & 63.5      \\
RPNet\cite{rpnet2021}            & -             & 68.2 \\ 
KPConv\cite{kpconv}            & 69.2             & 68.4 \\ 
FusionNet\cite{fusionnet2020}           & -             & 68.8 \\ 
JSENet\cite{jsenet2020}           & -             & 69.9 \\ 
RFCR\cite{rfcr2021}            & -               & 70.2 \\ 
CBL\cite{cbl2022}            & -             & 69.9 \\ 
SparseConvNet\cite{SparseConvNet2018}            & 69.3                & 72.5 \\ 
MinkowskiNet\cite{Minkowski2019}             & 72.2             & 73.6 \\ 
RepSurf-U\cite{repsurf2022}            &  -                  & 70.2     \\
Point Transformer\cite{pt2021} & 70.6               & - \\ 
Stratified Transformer\cite{strTransformer} & 74.3      & 73.7 \\
\midrule
\textbf{SAT}          &    \textbf{74.4}       & \textbf{74.2}      \\
\bottomrule
\end{tabular}
\end{center}
\caption{Quantitative results on ScanNetV2~\cite{datasetscannet} benchmark.}
\label{tbl-scannet}
\end{table}

\subsection{Ablation Study}\label{sec-4.3}
The following ablation experiments are conducted to study the impact of the proposed modules in SAT. The experiments are conducted on S3DIS, and the comparison results are shown in Table~\ref{tbl-ablation}.

Firstly, we remove the re-attention module. The improvement of the first row compared to the second demonstrates its effectiveness. We also find that Re-Attention significantly improves small classes of objects. Secondly, we remove the point-voxel shunted strategy in MGA and utilize summation instead to combine the features of the two branches. The improvement from the third to the second row confirms the shunted strategy's effectiveness. Finally, we remove PVCA in MGA and use ordinary point attention. Due to the lack of coarse-grained features and larger receptive field, the mIoU is reduced by 1.9\% compared to the second row.

\begin{table}
\begin{center}
\begin{tabular}{l|c}
\toprule
                         & mIoU (\%)            \\
\midrule
SAT         &   \textbf{72.6}        \\

removing Re-Atten. &   72.1          \\
\makecell[l]{removing Re-Atten. \& replacing shunted\\ strategy with summation}  &           71.5     \\
\makecell[l]{removing Re-Atten. \& replacing MGA\\  with ordinary point attention}  &       70.2         \\
\bottomrule
\end{tabular}
\end{center}
\caption{Ablation study on the SAT on S3DIS. }
\label{tbl-ablation}
\end{table}

\subsection{The Effectiveness of MGA}\label{sec-4.4}
The proposed MGA is a novel attention scheme for point cloud learning. It aggregates point and voxel tokens for each point based on the multi-head self attention. 

To show the effectiveness of the multi-granularity features provided by MGA, we replace the point attention in the swin3d transformer with a lite version of MGA. In the lite MGA, for a fair comparison, we keep the window size of the point attention branch and PVCA branch the same as that of the original swin3d transformer. The dimension of point attention becomes half of that in swin3d, and the number of voxels in PVCA is far smaller than points. Therefore, the lite MGA reduces the computational cost in attention operation by nearly half compared to the original swin3d.

As shown in Table~\ref{tbl-MGA}, swin3d with the lite MGA outperforms the original version by 0.8\% in mIoU with less computational cost. When using our standard version of MGA on S3DIS, the computational cost remains the same as Swin3d, but the performance is exceeded by 2.4\%.

It demonstrates that the multi-granularity features provided by MGA can also facilitate point cloud representation learning even without enlarger the receptive fields.

\begin{table}
\begin{center}
\begin{tabular}{l|c}
\toprule
Methods                         &  mIoU (\%)                   \\ 
\midrule
Swin3d Transformer         &   70.2          \\
replace point attention with lite MGA &   \textbf{71.0}          \\
\bottomrule
\end{tabular}
\end{center}
\caption{The validation of the effectiveness of the multi-granularity features provided by MGA.}
\label{tbl-MGA}
\end{table}

\subsection{The Analysis of Re-Attention Module}\label{sec-4.5}
\begin{figure}
\centering
\includegraphics[scale=0.5]{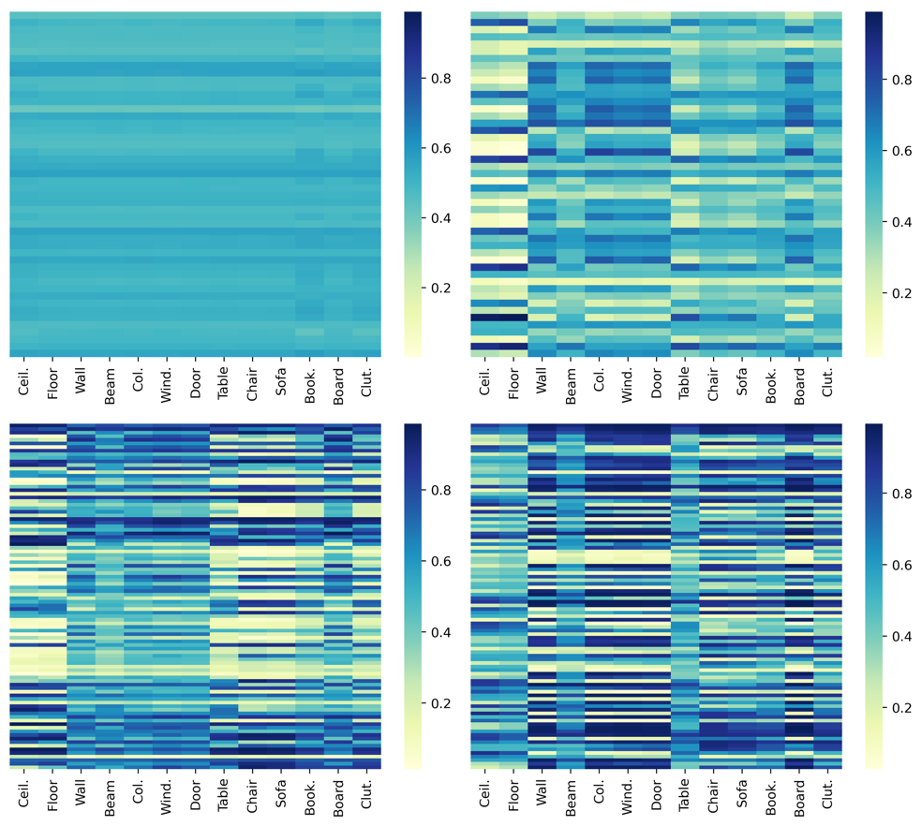}
\caption{Visualization of the output weights of Re-Attention module in different layers. \textbf{Top left}: first layer in stage 1; \textbf{Top right}: second layer in stage 1; \textbf{Bottom left}: first layer in stage 2; \textbf{Bottom right}: second layer in stage 2.}
\label{fig_heatmap}
\end{figure}

To further reveal how re-attention works, we visualize the output weights of points in different categories. We randomly select 30 scenes in S3DIS with more than 1.5 million points in the trained network. We compute the mean values of each dimension for the different categories and obtain the attention heat maps shown in~\Cref{fig_heatmap} in different layers.

In the first layer, the output values of re-attention for different categories and dimensions are almost indistinguishable. However, the different categories show differences starting from the second layer in stage 1. More excitingly, our re-attention module learned to find similarities between different categories (e.g., ceiling and floor, chair and sofa) and output a similar attention map. In the second stage, the maps have a more significant difference between dimensions, which indicates that re-attention plays a more significant role at deeper layers.

These maps demonstrate that our re-attention module can dynamically adjust the attention scores to the fine- and coarse-grained features output by MGA for each point according to the spatial and shape characteristics of their located objects.

\section{Conclusion}

In this paper, we propose the SAT to learn effective features in the scene point cloud with multiple-sized objects for semantic segmentation. The proposed methods are mainly based on two key components: the multi-granularity attention scheme and the re-attention module. The MGA efficiently aggregates tokens from large regions and disentangles the multi-scale features in the learning process. The re-attention module tailors attentive fields for objects of different sizes. Our SAT achieves state-of-the-art performances, and extensive experimental results demonstrate the effectiveness and efficiency of our methods. In addition, our work answers the critical question that it is feasible and effective to achieve object size-aware feature learning in point cloud segmentation.

\clearpage

{\small
\bibliographystyle{ieee_fullname}
\bibliography{main}
}

\end{document}